# The Semantic Brand Score

Fronzetti Colladon, A.

# The Semantic Brand Score

Fronzetti Colladon, A.

**Abstract**

The Semantic Brand Score (SBS) is a new measure of brand importance calculated on text data, combining methods of social network and semantic analysis. This metric is flexible as it can be used in different contexts and across products, markets and languages. It is applicable not only to brands, but also to multiple sets of words. The SBS, described together with its three dimensions of brand prevalence, diversity and connectivity, represents a contribution to the research on brand equity and on word co-occurrence networks. It can be used to support decision-making processes within companies; for example, it can be applied to forecast a company's stock price or to assess brand importance with respect to competitors. On the one side, the SBS relates to familiar constructs of brand equity, on the other, it offers new perspectives for effective strategic management of brands in the era of big data.

# 1. Introduction

Nowadays text data is ubiquitous and often freely accessible from multiple sources: examples are the well-known social media platforms Facebook and Twitter, thematic forums such as TripAdvisor, traditional media such as major newspapers and survey data collected by researchers. Consumers express their feelings and opinions with respect to products in multiple ways, and their attitude towards brands can often be inferred from social media (Fan, Che, & Chen, 2017; Mostafa, 2013). Consumers' interactions among themselves and with companies can influence prospective customers, firm performance and development of future products (e.g., Wang & Sengupta, 2016). The increase in availability of text data has raised the interest of many scholars who have been working towards the development of new automatized approaches to analyze large text corpora and extract meaning from them (Blei, 2012). At the same time, there has been significant interest in studying the value and importance of brands, considering both company and consumer-oriented definitions (Chatzipanagiotou, Veloutsou, & Christodoulides, 2016; de Oliveira, Silveira, & Luce, 2015; Keller, 2016; Pappu & Christodoulides, 2017; Wood, 2000). Customer-based brand equity was defined by Keller (1993, p. 1) as 'the differential effect of brand knowledge on consumer response to the marketing of the brand'; the author also presented brand image and awareness as the two main dimensions of brand knowledge. These dimensions are in many cases assessed using surveys, case studies, interviews and/or focus groups (Aaker, 1996; Keller, 1993; Lassar, Mittal, & Sharma, 1995). Such approaches can be time-consuming for large samples and are sometimes biased, due to the fact that consumers often know to be observed and studied (making their expressions less natural and spontaneous). Another problem of past models is that brand equity dimensions are often many, heterogeneous and sometimes not easy to integrate in the final assessment.

Among many factors affecting consumer-based brand equity, attention paid to consumers' feedback has proved to play a major role (Battistoni, Fronzetti Colladon, & Mercorelli, 2013). Therefore, in the era of big data, it seems relevant to investigate the opinions of consumers and other stakeholders in their spontaneous expressions – while, for instance, discussing the characteristics of a product, or their user experience, without them having the perception of being monitored. Nowadays, social media and online reviews represent a common method of feedback. However, dealing with very large datasets usually requires rapid and automatized assessments that would be unfeasible when relying on traditional surveys.

This work presents a new measure of brand importance – the Semantic Brand Score (SBS) – which overcomes some of these limitations, being automatable and relatively fast to compute even on big text data, without the need to administer surveys or to inform those who generate contents (such as social media users). The Semantic Brand Score (SBS) can be calculated for any set of text documents, either customer-based or related to the opinions and experience of other stakeholders of a company; it can be applied to different contexts: newspapers, social media platforms, consumers' interviews, etc... Indeed, a good measure of brand importance should be sensitive to its variations and should be applicable across markets, products and brands (Aaker, 1996).

In this work, brand importance is computed based on text data and conceptualized as the extent to which a brand name is utilized, it is rich in heterogeneous textual associations and 'embedded' deeply at the core of a discourse. Accordingly, the SBS is expressed along the three dimensions of brand prevalence, diversity and connectivity, as illustrated in the next sections. Even if the way the SBS measures brand importance is new, it partially reconnects to dimensions discussed in other well-known models, such as brand awareness and heterogeneity in brand associations (Aaker, 1996; Grohs, Raies, Koll, & Mühlbacher, 2016; Keller, 1993). This approach is also an attempt to reduce the gap between text analysis and the study of brand importance, as this research area remains mostly unexplored even when considering well-known text statistics – such as the study of word co-occurrences or term frequencies (Evert, 2005).

The calculation of the SBS combines methods of Social Network and Semantic Analysis (Wasserman & Faust, 1994), using word co-occurrence networks (Danowski, 2009; Leydesdorff & Welbers, 2011). This paper advances research in this direction and can be useful for brand managers who would like to monitor and improve the equity of their brands and products. Additionally, the SBS is proposed as an adaptable metric, which can be applied to different sets of words – not just brands – with the possibility of multiple uses (such as the study of the strength of keywords associated with the main core values of a company).

## 2. Measuring Brand Importance

Customer-based brand equity was traditionally conceptualized along the dimensions of brand awareness and brand image (Keller, 1993); the former referring to brand recall and recognition, the latter to brand associations, their uniqueness, type and strength. The dimensions of brand knowledge were discussed in a subsequent study by Keller (2003) which demonstrated that marketing activities can generate feelings, thoughts, attitudes and experiences that can influence consumers' response and purchase intentions. Grohs et al. (2016) focused on brand associations, investigating the impact of their number, uniqueness, consensus and favorability on brand strength – finding a positive effect for perceived consensus, size and favorability. Another important model, based on information economics and signaling theory was proposed by Erdem and Swait (1998). In the authors' work, brands were seen as the company's response to customers' uncertainty and information asymmetries, ultimately serving as a signal of product quality. The authors' framework included factors such as brand investments, credibility, perceived quality and information costs saved. Other studies attributed greater importance to consumer experiences, satisfaction and brand loyalty (Nam, Ekinci, & Whyatt, 2011). Christodoulides and de Chernatony (2010) published a review which distinguished between financial based measures of brand equity and consumer-based perspectives, with the latter comprising both direct approaches focused on the evaluation of consumers' preferences and indirect approaches centered on the analysis of outcome variables, such as the price premium.

Measurement of brand equity was usually developed around market surveys which were administered to consumers and other potential stakeholders, or based on financial methods, in some cases considering the differential value of a product with and without its brand (i.e. the price premium). Lassar et al. (1995) produced several survey items to assess customer-based brand equity, organized in the dimensions of performance, social image, value, trustworthiness and attachment. Aaker (1996) illustrated five sets of measures to evaluate brand equity – loyalty, perceived quality/leadership, awareness, association/differentiation and market behavior – and presented the items useful to assess them. The author claimed that price premium was one of the best measures in most contexts. Price premium was given great importance in more recent studies as well (Netemeyer et al., 2004). However, this indicator offers only a partial view and cannot be used in those contexts where sales and profits are not among the objectives of a company or organization. Battistoni, Fronzetti Colladon and Mercorelli (2013) used the Analytic Hierarchy Process to rank factors that could influence customers' perceptions about a brand, thus affecting both the brand image and its awareness. Their research proved the importance of monitoring and maintaining a successful dialogue with customers paying particular attention to the received feedback. Indeed, attention paid to customers' feedback turned out to be the most important factor, immediately after the company history and reputation. In this sense, the text analysis of consumers' interactions on social media can be of great importance (Malthouse, Haenlein, Skiera, Wege, & Zhang, 2013).

Past research frequently addressed the effect of social media on brand equity, using different approaches and considering different points of view. Hollebeek, Glynn and Brodie (2014) conceptualized consumer brand engagement on social media and developed a measurement scale, based on survey questions. Other scholars provided evidence to suggest the positive impact of social media marketing activities on brand equity (Bruhn, Schoenmueller, & Schäfer, 2012; A. J. Kim & Ko, 2012). Laroche, Habibi, Richard and Sankaranarayanan (2012) showed that online brand communities can enhance brand loyalty. Even though all these studies investigated the link between social media activities and brand equity, their measurement of brand related constructs always relied on survey questions. On the other hand, it seems important, at least in online contexts, to find a measure which can be directly inferred from analyzing the discourse of social media users – trying not to impact their spontaneous behavior and without asking them to complete a survey.

The approach presented in this paper goes in this direction, without the aim of directly producing a single score representing brand equity as the expression of a positive construct. The author proposes a new measure of brand importance, based on the analysis of the occurrences of a brand name in a discourse, its embeddedness in text data, and the heterogeneity of its text associations. Brand importance is conceptualized by using the three dimensions of brand prevalence, diversity and connectivity (described in Section 2.1). According to this approach, a brand that is used marginally, or that is very peripheral in a set of documents, is classified as unimportant. An important brand, on the other hand, is at the core of a discourse, with the possibility of being associated to either negative or positive feelings. Therefore, a more comprehensive picture regarding the value of a brand is obtained combining the SBS with sentiment analysis, as illustrated in Section 3.2. The approach presented here is new and not necessarily limited to the analysis of customers' expressions, even if it is partially linked to some dimensions of the brand knowledge model presented by Keller (1993). Indeed, the conceptualization of brand importance presented in this paper is relevant to brand equity, even if it does not automatically translate into it. Keller's (1993) definition of brand equity includes the concept of differential response to knowledge of a brand name, which suggests that knowing the brand name is the main starting point. This knowledge is captured by two dimensions of the SBS, prevalence and connectivity, which not only reflect the brand name frequency of use, but also its embeddedness at the core of a discourse. In addition, the differential response may originate from the brand associations, in relation to their number and valence, with the former captured by the SBS dimension of diversity. While valence of associations can point to the nature of the differential response, the basis for a differential response is the importance of a brand. As described in the next section, this is captured by the SBS through the measures of prevalence, diversity and connectivity, whereas a measure of valence of brand associations, although still relevant, is not directly included in the SBS for the reasons illustrated in Section 3.2.

In online contexts, some efforts in the direction of evaluating brand popularity were made by counting the number of likes to brand pages and the number of comments on Facebook (De Vries, Gensler, & Leeflang, 2012). Gloor (2017) developed a software tool (Condor) useful for web and email data collection, and the calculation of social network and semantic metrics. The author used these metrics to support the idea that important brands are often associated with high levels of brand-related interaction on social media. Even if the value provided by these assessments is remarkable, they only seem to focus on specific aspects of brand equity, such as brand popularity (Gloor, Krauss, Nann, Fischbach, & Schoder, 2009). Moreover, these methodologies are often applied to web data and are based on the study of social interaction, or links among webpages or user generated contents (Yun & Gloor, 2015); they do not consider word co-occurrence networks and might be only partially applicable in contexts were text documents are not associated to an interaction network – for example, a press review. Very few other studies tried to link textual analysis with factors affecting brand equity. Aggarwal, Vaidyanathan and Venkatesh (2009) studied online brand representations, extracting information from the Google search engine; they evaluated brand positioning by examining the association between brands and various adjectives or descriptors. Gloor et al. (2009) proposed to look at the centrality of brands in semantic networks extracted from the web, considering centrality as a proxy for brand popularity.

Overall, many different models and measurement approaches have been developed in previous studies. However, it appears that just few of them attempted to use text statistics for brand management and that a more comprehensive measure which can be utilized to assess brand importance from big text data – integrating different components and not only focusing on a single construct – is still missing. This is in part surprising given the large influence that social media communication can have on brand loyalty and equity creation (Bruhn et al., 2012; Erdoğmuş & Çiçek, 2012). Moreover, many practitioners found past methodologies difficult to apply and understand (Christodoulides & De Chernatony, 2010). To solve some of these problems, the Semantic Brand Score, presented in the next paragraph, might be of help; it supports, for example, the assessment and monitoring of customer perceptions in relation to a brand as expressed on social media, or the evaluation of brand importance in newspaper articles. On the whole, the SBS can assist the analysis of brand positioning in any text data (which might come from customers, competitors, financial analysts, or other stakeholders of a firm).

### 2.1. Dimensions of the Semantic Brand Score

The Semantic Brand Score has been conceived with the aim of allowing a rapid assessment of brand importance while considering multiple points of view and opinions included in relatively large sets of text documents. Its calculation combines methods of semantic and social network analysis (Leydesdorff & Welbers, 2011; Wasserman & Faust, 1994) and is based on the construction of word co-occurrence networks (where words that co-occur within a specific range are linked together and represented as nodes in a graph). This metric can be calculated for any text corpora and potentially be used across different languages, cultures, knowledge domains and research settings. In this paper, the SBS is presented as the combination of the three dimensions of prevalence, diversity and connectivity (see Figure 1). Even if these dimensions represent new constructs, their conceptualization was inspired by important components of widely accepted brand equity models (Keller, 1993; Wood, 2000) and by common statistics in text analysis – such as term frequency and the study of word co-occurrences (Evert, 2005). A more detailed discussion of these connections is provided in the following sections.

Prevalence represents the frequency with which the brand name appears in a set of text documents. The more a brand is mentioned, the higher its prevalence, with higher expectation of familiarity for the text authors. A more recurrent brand name is probably more visible and easier to recall than a brand that appears rarely. On social media, for example, only users who are aware of a brand name, and can remember it, will use it. Moreover, readers will probably memorize more easily a brand that is frequently repeated, with respect to one that rarely appears in a discourse. This suggests a possible link between prevalence and the dimension of brand awareness (Aaker, 1996; Keller, 1993), which is however only partial as there might be brands with high awareness that are not frequently mentioned in text documents, depending on their specific market.

Diversity is partially linked to the concept of lexical diversity (McCarthy & Jarvis, 2010) and to the study of word co-occurrences (Evert, 2005); it measures the heterogeneity of the words co-occurring with a brand. This measure is connected to the construct of brand image and therefore to the association of other words with the brand (Keller, 1993; Wood, 2000). While the interpretation of favorability, uniqueness and type of these associations is important (Grohs et al., 2016) – and possible since supported by the construction of the co-occurrence networks illustrated in Section 3 –, diversity focuses on their heterogeneity. A brand could be mentioned frequently in a discourse, thus having a high prevalence, but always used in conjunction with the same words, being limited to a very specific context. On the other hand, a brand with more diverse associations should be preferred as this is usually embedded in a richer discourse, with higher versatility of the brand name. The metric of diversity is supported by past research showing that the number of brand associations positively affects brand strength (Grohs et al., 2016). Ultimately, the presence of diverse lexical associations to persons, things or other concepts could increase brand awareness (Aaker, 1996; Keller, 2016). As better discussed in Section 3.2, each dimension of the SBS – in particular diversity – has to be interpreted and commented considering the sentiment, strength and type of associations.

Connectivity is a new measure in the brand equity scenario, derived from social network analysis and operationalized through the betweenness centrality metric (Wasserman & Faust, 1994). As better described in Section 3, this measure indicates how frequently a node in a graph lies in-between the shortest paths which interconnect the other nodes. When nodes represent words, they express how often a word (brand) serves as an indirect link between all the other pairs of words. Connectivity reflects the embeddedness of a brand in the co-occurrence network, going beyond the direct connections considered in the calculation of diversity. One could imagine connectivity as a measure of 'how much of a discourse passes through a brand', i.e. how much a brand can support a connection between words which are not directly co-occurring. A high connectivity level could indicate a brand which serves as a bridge between different sets of words. Thinking more laterally, connectivity could be considered as the "brokerage" power of a brand, i.e. its ability of being in-between different groups of words, sometimes representing specific discourse topics. This measure differentiates from diversity: it might be the case that a brand has high diversity, being surrounded by many heterogeneous terms, but if peripheral still presents low connectivity, since not properly integrated in the general discourse. Connectivity has already proved its value in previous research: the betweenness of a company name in a text co-occurrence network was used by Fronzetti Colladon and Scettri (2017) as predictor of a company stock price (the analysis was performed based on the company employees' discourses on an intranet social network). In addition, Gloor et al. (2009) found that the betweenness centrality of a brand, in semantic networks extracted from the web, could be considered as a proxy for its popularity.

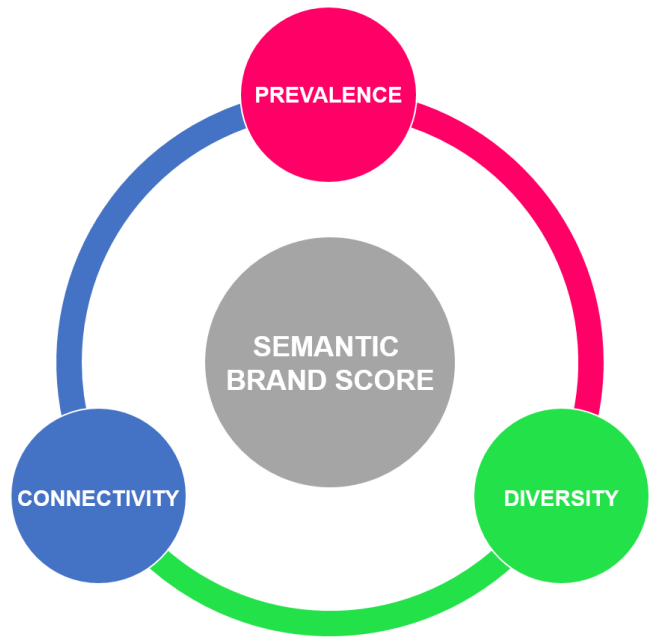

**Figure 1.** Dimensions of the Semantic Brand Score.

### 3. Calculation of the Semantic Brand Score

The Semantic Brand Score is a measure of the importance of a brand in a set of text documents. For example, the set of documents could include the transcripts of interviews to consumers, the more spontaneous online posts by company stakeholders on Facebook or Twitter and the news and comments posted by employees on an intranet social network. Once the set of documents has been identified (and collected), the analyst should proceed with some common text preprocessing procedures, as described in the book about the NLTK package of the programming language Python 3 (Perkins, 2014). The most common steps at this stage are: (a) the removal of html tags, hashtags, punctuation, special characters, numbers and stop-words (i.e. words like "the" and "a" which usually provide little contribution to the meaning of a sentence);

(b) the transformation of all the text into lower case words; (c) the computation of bigrams (pairs of words recurring together in the same order); (d) the extraction of stems, removing the affixes of words (Jivani, 2011), or the replacement of words with their root words using a dictionary – also known as lemmatization (Asghar, Khan, Ahmad, & Kundi, 2014). These steps are necessary to reduce the language complexity and retain the most significant words, which can better convey the meaning of a discourse. To give an example, the following sentence:

*"Aurora was bright as the DAWN and yet she was mysterious and dark, as the night surrounding the stars."*

could be transformed into a list of tokenized and stemmed words as the following – note that NLTK Snowball Stemmer has been applied here (Perkins, 2014), see for example how the word "mystery" has been changed:

[aurora, bright, dawn, yet, mysteri, dark, night, surround, star]

In general, one should always adapt the text preprocessing procedures to the specific context that is being analyzed, in order to avoid removing automatically words which could be relevant for a specific study: for example, if the brand is a number, one should avoid removing numbers. Similarly, the analyst should choose whether it is appropriate to remove verbs (such as 'to be' and 'to have'), and be careful with the lemmatization of words, because important acronyms or distinctive terms could be lost during this process. For text preprocessing the analyst could use the programming language Python with the library NLTK (Perkins, 2014) or other specific packages and functions written using the programming language R, such as the "textProcessor" included in the STM package (Roberts, Stewart, & Tingley, 2015). However, other options are also possible and the calculation of the Semantic Brand Score is not constrained to the use of a specific software (it can be programmed in any language).

Once the original documents have been preprocessed, the calculation of the SBS passes through their transformation into a word co-occurrence network (Bullinaria & Levy, 2012; Dagan, Marcus, & Markovitch, 1995; Danowski, 2009; Diesner, 2013). This network is made of *n* nodes G = {$g_1$, $g_2$, $g_3$, ... $g_n$} – each one representing a word – and *m* arcs; the arc $a_{ij}$, which originates at node *i* and terminates at node *j*, exists if the term *i* precedes the term *j* in at least one text document, within a range of five words. Five is a threshold chosen as it led to good results in previous research (e.g., Fronzetti Colladon & Scettri, 2017); however, the analyst is free to change this number to one that is appropriate for his/her analysis. Tests to assess the effects produced by a different threshold choice are presented in Section 4: they show that changing the co-occurrence range does not affect the relative ranking between brands (i.e. if the SBS of Brand A is higher than the one of B with a threshold of five words, this ranking is confirmed also for higher or lower thresholds). Nonetheless, these preliminary findings should be further explored and commented in future research. Words that co-occur more than once will produce multiple arcs. Before the computation of the score of each SBS dimension, the network is symmetrized – i.e. directed arcs are replaced with undirected edges. Moreover, loops are removed and multiple arcs between two nodes are substituted by a single edge weighted with a value equal to the number of multiple arcs. Another option for the analyst is to remove arcs with low weights which indicate rare co-occurrences. Figure 2 shows an example of the network which would be generated if considering the following two documents (represented as two lists of preprocessed words):

[[aurora, bright, dawn, yet, mysteri, dark, night, surround, star],

[unexpect, mysteri, dream, sometim, come, true, dark, night]]

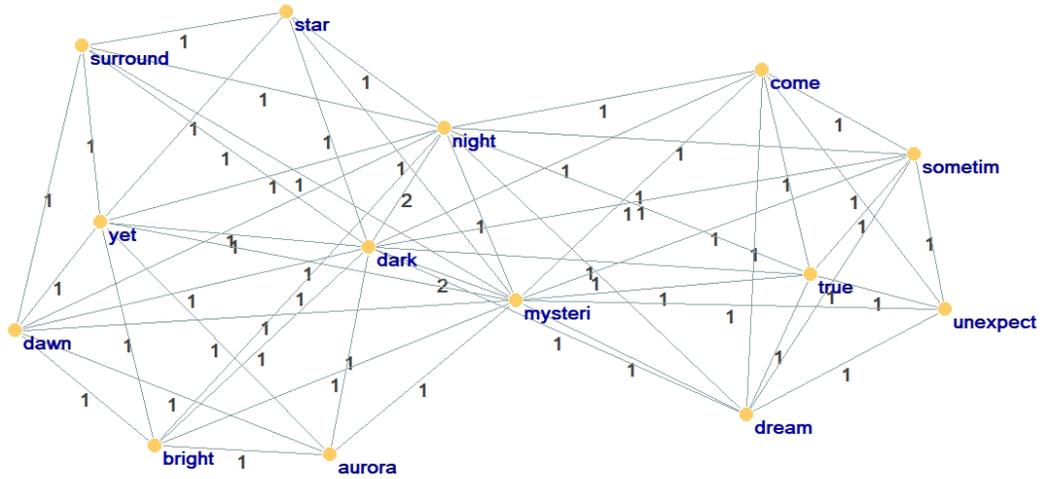

**Figure 2.** Example of word co-occurrence network.

### 3.1. Prevalence, Diversity and Connectivity

The SBS has been conceived to express the importance of a brand along the three dimensions of prevalence, diversity and connectivity, which come from the combination and novel use of well-known social network and semantic metrics (Evert, 2005; Wasserman & Faust, 1994).

*Prevalence.* Prevalence refers to the presence of a specific word (brand) in the text and is calculated as its overall term frequency $f(g_i)$, for a given set of documents and/or timeframe. In order to compare this value across different text corpora of different sizes, one could divide the absolute term frequency by the total number of words in the text (indicated as *totW*).

$$PREV(g_i) = f(g_i)$$

$$PREV'(g_i) = \frac{f(g_i)}{totW}$$

To calculate prevalence, one does not need to transform the documents into a word co-occurrence network, as described at the beginning of Section 3.

*Diversity.* Diversity expresses the heterogeneity of the words surrounding a brand. It answers the question "how many different network neighbors does the brand have?". The higher the diversity, the more heterogeneous the semantic context in which the brand name is used. Diversity is calculated as the *degree centrality* of the brand in the co-occurrence network (Freeman, 1979), i.e. the number of edges directly connected to its node. This number is higher when a brand co-occurs with many different words. By contrast, if the discourse around a brand converges towards a small set of words, its diversity will be lower. In the following formula *d(g_i)* indicates the degree of the brand node *g_i*.

$$DIV(g_i) = d(g_i)$$

*Lexical diversity* is a different but similar measure which was explored in past research (McCarthy & Jarvis, 2010) to indicate the number of different words used in a text, over the total number of words. However, this measure usually refers to an entire text body and not to the terms used in the close range of a specific word. Moreover, lexical diversity has been criticized as it can be very sensitive to variations in text length (Malvern, Richards, Chipere, & Durán, 2004).

In order to allow a comparison between networks of different sizes, diversity can be divided by the maximum value of degree centrality, which is *n – 1* with *n* equal to the total number of nodes (Wasserman & Faust, 1994):

$$DIV'(g_i) = \frac{d(g_i)}{n-1}$$

A network metric which is worth discussing as somehow in-between prevalence and diversity is the *weighted degree centrality* of the node representing the brand. This measure, well described in the work of Wasserman and Faust (1994), sums the weights of the edges which are adjacent to a specific node. Consistently, its score for a brand node is represented by the total number of text co-occurrences (repeated co-occurrences are counted more than once). While weighted degree centrality seems to be interesting in this context, the author still suggests referring to term frequency to evaluate prevalence. Indeed, based on the network construction technique just described, weighted degree centrality would have the disadvantage of favoring the words that are used in the middle of a text document against those used at its extremes (since central terms usually have more network neighbors given that co-occurrences are calculated considering a sliding window range). With regard to diversity, using weighted degree centrality in lieu of simple degree would extend the original measure with a proxy for the strength of each association (edge weight) with directly connected nodes. However, this would have the drawback of disconnecting diversity from the concept of heterogeneity, as it could lead to a very high score only because a single connection is repeated multiple times (instead of having multiple links to different words). On the other hand, a problem of diversity could lie in the fact that a brand node could have many different direct connections which occur very rarely (indicating probably insignificant associations). In such a scenario, it is often more convenient to filter out the edges representing very rare co-occurrences – such as the words co-occurring just once or twice, also taking into account the dataset size.

*Connectivity.* Connectivity depends on the network position of a brand. Its value is higher when the brand node is more often in-between the network paths which interconnect the other words within the text and therefore more deeply embedded in a discourse. Connectivity answers the question "how often does a brand figure in the network paths that keep together other pairs of words (or other parts of the discourse)?". Connectivity has been operationalized with the measure of *betweenness centrality* (Freeman, 1979), which is widely used in social network analysis as a measure of brokerage, influence or control of information that goes beyond direct links (Wasserman & Faust, 1994). Connectivity of the brand $g_i$ is given by the formula:

$$CON(g_i) = \sum_{j<k} \frac{d_{jk}(g_i)}{d_{jk}}$$

with $d_{jk}$ equal to the number of the shortest paths linking the generic pair of nodes $g_j$ and $g_k$, and $d_{jk}(g_i)$ to the number of those paths which contain the node $g_i$. Applying a similar approach to the one used for diversity, it can be useful to normalize connectivity to allow a comparison between networks of different sizes. This can be done dividing the value of connectivity by [(n – 1) (n – 2) / 2], which is the total number of pairs of nodes not including $g_i$ (Wasserman & Faust, 1994):

$$CON'(g_i) = \sum_{j<k} \frac{d_{jk}(g_i)}{d_{jk}} \Big/ [(n-1)\ (n-2)\ /\ 2]$$

*Semantic Brand Score.* The Semantic Brand Score is obtained by summing the standardized values of prevalence, diversity and connectivity, in order to attribute the same importance to each dimension, even in the case of different variances.

$$SBS(g_i) = \frac{PREV(g_i) - \overline{PREV}}{std(PREV)} + \frac{DIV(g_i) - \overline{DIV}}{std(DIV)} + \frac{CON(g_i) - \overline{CON}}{std(CON)}$$

In the above formula, each component has the same weight, as the three dimensions are all considered to carry the same importance. However, the reader should note that future studies could explore the possibility of weighting this sum with unequal coefficients.

The measures of prevalence, connectivity, diversity and the SBS could be calculated applying a min-max normalization, or considering percentile rank scores. This could be sometimes useful for comparative purposes within a specific data sample, and would be aligned with the idea that the SBS has a meaning that should always be discussed with a specific reference to the context and set of documents which are being analyzed. To put it in other words, one could assess and compare the SBS of one brand on Twitter and on Facebook and unsurprisingly obtain different values; apart from the final numbers, what is often more interesting is the difference between competing brands and the evolution of their SBS over time.

### 3.2. Sentiment, Associations and Major Topics

It is important to notice that the Semantic Brand Score does not tell us if the lexical associations around a brand are positive or negative. A brand could be frequently mentioned and very central in a discourse, surrounded by heterogeneous terms, but the same discourse could express negative feelings. For this reason, the SBS should always be interpreted together with the characteristics of the specific context where it is measured. In the hypothetical case just described, or during a crisis, business managers could desire a lower SBS for their company name on social media. To evaluate the positivity or negativity of a set of text documents, sentiment analysis might come to help. Many different software programs or programming languages allow a multilingual classification of the sentiment of text documents (Gloor, 2017; Mostafa, 2013). One could also expand the sentiment analysis with the evaluation of other dimensions of the language used, such as complexity and emotionality (Alonso, Cabrera, Medina, & Travieso, 2015; Gloor, 2017). Accordingly, the analyst could classify and comment the text documents which include the brand name. In addition, it would be possible to split the text corpora into two subsets, one with positive and the other with negative posts/statements about the brand. In this way, the SBS could be calculated twice, once for each subset, to determine where the brand is stronger.

Studying the sentiment of the words surrounding a brand in the co-occurrence network would be another possibility. In particular, one could study their polarity (positivity or negativity) inferring their sense from other words in the context (Basile & Nissim, 2013). This could be an interesting choice when each text document is significantly long and the brand is mentioned rarely (with the possibility, for instance, of having an average negative sentiment for the entire document, but a positive sentiment in the phrases where the brand is mentioned). However, this last approach could produce the disadvantage of ignoring the sense of entire phrases which include the brand, where instead the order and combination of words can be important to assess correctly their meaning. Assuming the polarity of the brand co-occurrences is correctly calculated – and expressed in the range [-1,1], where negative numbers indicate, on average, a negative sentiment and positive numbers a positive sentiment –, one could consider the possibility of multiplying this score by the SBS, in order to obtain a final number which takes into account the valence of the textual brand associations. Nonetheless, a significant amount of information would be lost if only looking at this final number: a score of zero might come either from a very high SBS and a totally neutral sentiment, or just from a brand which is never mentioned and therefore has a zero SBS. As a result, the suggestion is to have two separate scores (polarity of associations and SBS) which could be interpreted together.

Lastly, one could isolate a set of positive and negative words and evaluate their SBS to assess their influencing power. The SBS, indeed, is a measure which can be applied to any word and not just to brands; therefore, it is flexible and available for multiple uses. For example, one could consider the keywords associated to a set of core values of a company and study their importance. A possible step, depending on the analysis, could also be to merge the nodes representing each core value, before calculating their diversity and connectivity.

In addition to sentiment analysis, the analyst could model the main topics in the discourse about the brand. In this regard, probabilistic topic modeling algorithms might come to help, as a convenient way to extract meaning from large text corpora, which would otherwise require a lot of time to be analyzed by human readers (Blei, 2012). A preliminary idea about the main topics related to a brand can also be obtained from the analysis of the network neighbors of the brand node, i.e. looking at the words that most frequently co-occur with the brand. Thanks to the construction of the co-occurrences network, the analyst is provided with an immediate representation of the possible brand associations in the text and their strength (see Figure 2). Therefore, potentially it could be useful to count – and/or categorize – those favorable against those negative (Grohs et al., 2016), with the possibility of weighting this count by the individual strength of each brand association (represented by the weight of the edge connecting each word to the brand node). Similarly, one could investigate and compare the uniqueness of each association, when multiple brands are considered in the analysis (from a network perspective, this would correspond to looking for the direct connections that are not shared among different brand nodes).

The considerations made in this paragraph can help to reconcile the SBS with the construct of brand image (Keller, 1993); this section stresses the importance of discussing the SBS value together with the uniqueness, strength and positive or negative sentiment of the words surrounding a brand. To put it in other words, a high SBS is usually associated with a brand that is named frequently and that appears at the hearth of a discourse. However, this could convey either the appreciation or dislike of text authors (for example consumers interacting on Facebook), as the brand could be mentioned in relation to something bad or good. This is the reason why the analyst can achieve a more appropriate indication of the value of a brand if he/she interprets the SBS together with the positivity or negativity of brand co-occurrences. Moreover, the sentiment of the brand co-occurrences and the SBS provide company managers with different information that potentially call for different types of action. For example, one could consider a brand which values visibility on social media. A low SBS on Twitter would suggests some form of intervention would be required to increase the brand presence on this platform. A negative sentiment of the brand co-occurrences, on the other hand, could motivate an intervention meant to improve the dialogue with consumers, in order to reduce the causes of discontent. In general, if a brand is almost never mentioned in a specific context, this will result in a low SBS, suggesting a non-pivotal role of that brand, regardless of the valence of brand co-occurrences. Finally, it is important to notice that brand co-occurrences are intended as a possible proxy for brand associations in the text authors' minds, but not as a rigorous measure of them.

## 4. Possible Applications

SBS can be calculated for any peculiar word or set of words in a set of documents. This measure is very flexible and can be used for several purposes – including the evaluation of the internal and external importance of a brand, performed by analyzing its SBS in the discourse of a company's employees interacting on an intranet social network, or in the dialogue of a company's stakeholders on Twitter. Even if the discussion of these possibilities should be addressed by dedicated research, some preliminary experiments and applications are briefly introduced in this section.

Figure 3 shows the evolution of the SBS over time, calculated for two major brands of a large multinational company. The calculation was carried out considering the interactions of more than 10,000 employees on the company intranet forum, from September 2015 to April 2017. Employees posted news and commented on the news posted by others on company-related topics (as per agreed privacy arrangements, more details about the company or the platform used cannot be disclosed). Earlier in 2015, the company decided to abandon its historical name in favor of a new brand (already owned and partially used). This initiative was supported by big internal and external advertising campaigns and by internal communication activities which involved all the employees. As the graphs shows, the old brand was already weaker in 2015 (as the transition process had already started). What is even more interesting is to look at the time evolution of the SBS which increases for the new brand and decreases for the old one (time is represented by the changing colors in the three-dimensional scatter plot). This is a proof of the successful communication strategies of the company. These results, obtained from the analysis of the SBS on the intranet social network, are aligned with those obtained by a well-known external research institute, hired by the company. The institute measured the equity of the two brands, by means of more traditional surveys (Aaker, 1996; Keller, 1993). These surveys assessed brand reputation and awareness among business clients and consumers, using questions such as "When you think about computer games what is the first brand that comes to your mind?" ('computer games' is just an example here). The rankings obtained from the surveys are consistent with the time trends of the SBS for the two brands (old and new), as measured on the company intranet social network. This link could be explored further in future research with the objective to test whether the SBS could be used as a proxy for some of the constructs mapped by traditional brand equity surveys.

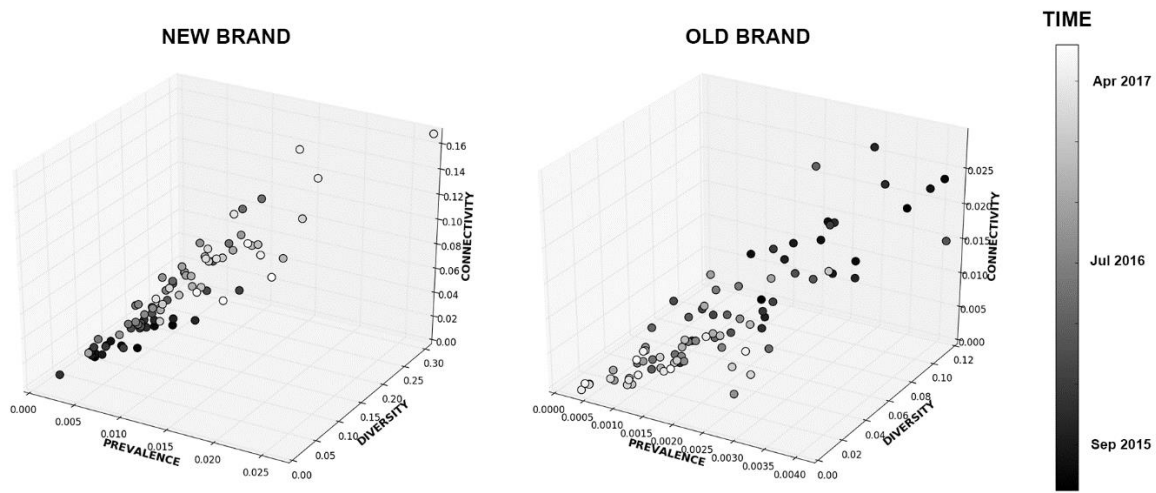

**Figure 3.** Evolution of the Semantic Brand Score during a brand transitioning.

The second step of the analysis was to perform Granger-causality tests to see if the evolution of the SBS and its components could precede and potentially help to predict variations in the company stock price (a stationary variable calculated as the weakly difference in price). This exploratory choice was inspired by past research which studied the internal communication of the employees working for a large company, using text analysis to identify measures that could help to predict the company stock price (Fronzetti Colladon & Scettri, 2017). The authors demonstrated the significance of metrics which were very similar to connectivity and prevalence. Consistently, as shown in Table 1, all the components of the SBS were tested as possible predictors (up to six-week lags), considering a time period of 85 weeks.

| Dependent: Stock Price | $\chi^2$ Lag 1 | $\chi^2$ Lag 2 | $\chi^2$ Lag 3 | $\chi^2$ Lag 4 | $\chi^2$ Lag 5 | $\chi^2$ Lag 6 |
|---|---|---|---|---|---|---|
| Prevalence | .51 | .95 | 9.78* | 11.51* | 11.28* | 11.73 |
| Diversity | .78 | 1.25 | 10.27* | 12.13* | 11.37* | 11.61 |
| Connectivity | 1.08 | 1.52 | 11.36** | 13.41** | 12.32* | 12.10 |
| Semantic Brand Score | .77 | 1.22 | 10.46* | 12.30* | 11.59* | 11.79 |

*$p < .05$; **$p < .01$.

**Table 1.** Granger causality tests.

As Table 1 shows, both the SBS and all its components can significantly granger-cause the weakly price variation at different lags and could, for instance, be tested in an ARIMAX model (Box, Jenkins, & Reinsel, 2013) for forecasting purposes. Connectivity is the most important predictor in this example. Significant results are for lags 3 to 5, suggesting that internal communication dynamics can help to forecast the company stock price, three to five weeks in advance.

In some specific contexts, companies might be interested to assess the importance of their brand and compare it with the one of their competitors. For example, companies could collect and analyze the social media posts about a set of brands or about specific topics (identified by a list of keywords), to check whether their brand name is predominant and well-embedded in the discourse. An example is provided in Figure 4, where the author collected, for a period of four days in July 2017, the Italian Twitter discourse with regard to three major telecommunication companies in Italy. Data was collected using the social network and semantic analysis software Condor (Gloor, 2017). In Figure 4, each network node represents a Twitter user (except for the three squared nodes which represent the search queries) and each network link is the expression of a tweet, a retweet, an answer or a mention. The sentiment was positive for all the three brands. From the picture – and from some first analytics implemented in Condor – it was immediately clear that Brand C was the strongest in this context: its debate was more vivid with response times lower on average, it had predominant occurrences, its network was denser and the posts involving the brand name attracted more participants. The SBS of the three brands (also shown in the picture) provided consistent results and could help to quantify the distance between these brands.

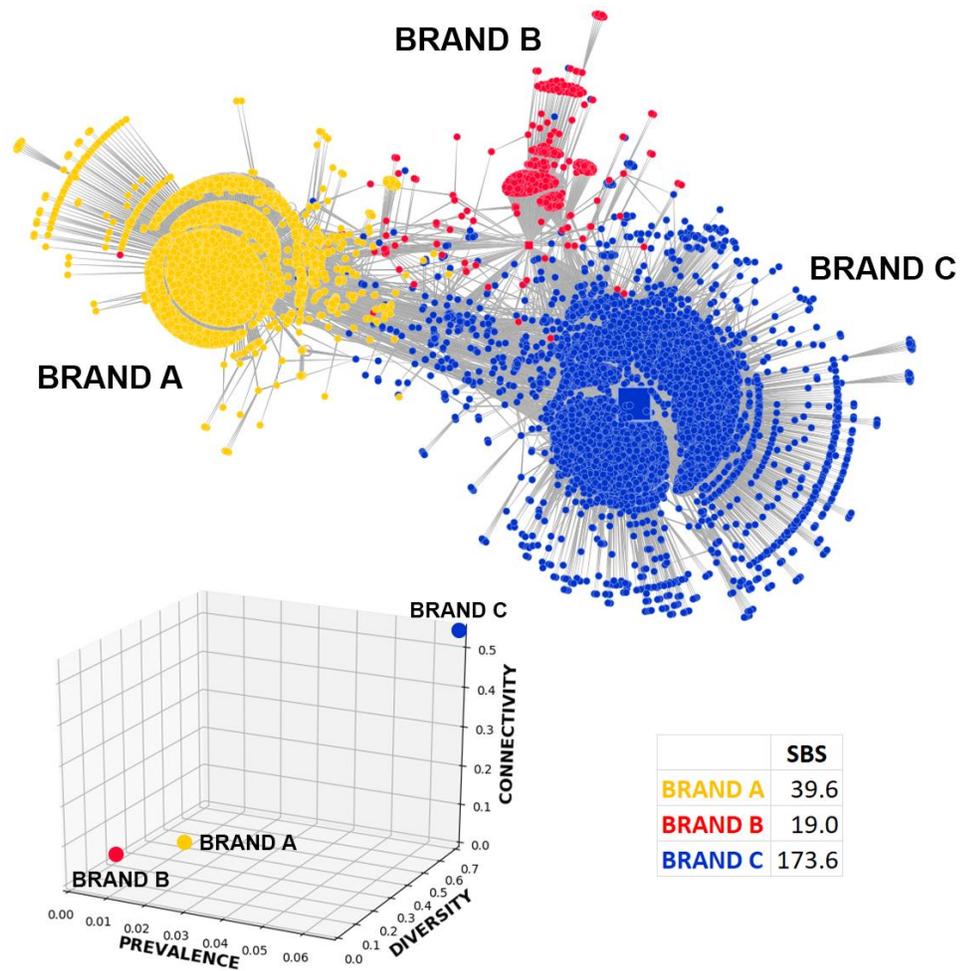

**Figure 4.** Comparing the SBS of multiple brands on Twitter.

One additional point of interest is to understand if variations of the word co-occurrence range chosen to calculate the SBS can influence the analysis. To this purpose, the author replicated the two experiments described above, and calculated the SBS – for the old and new brand in the intranet setting (considering the third quarter of 2015) and for the three brands in the Twitter experiment – testing different word co-occurrence thresholds (from 1 to 20 words). Results of these tests are presented in Figure 5.

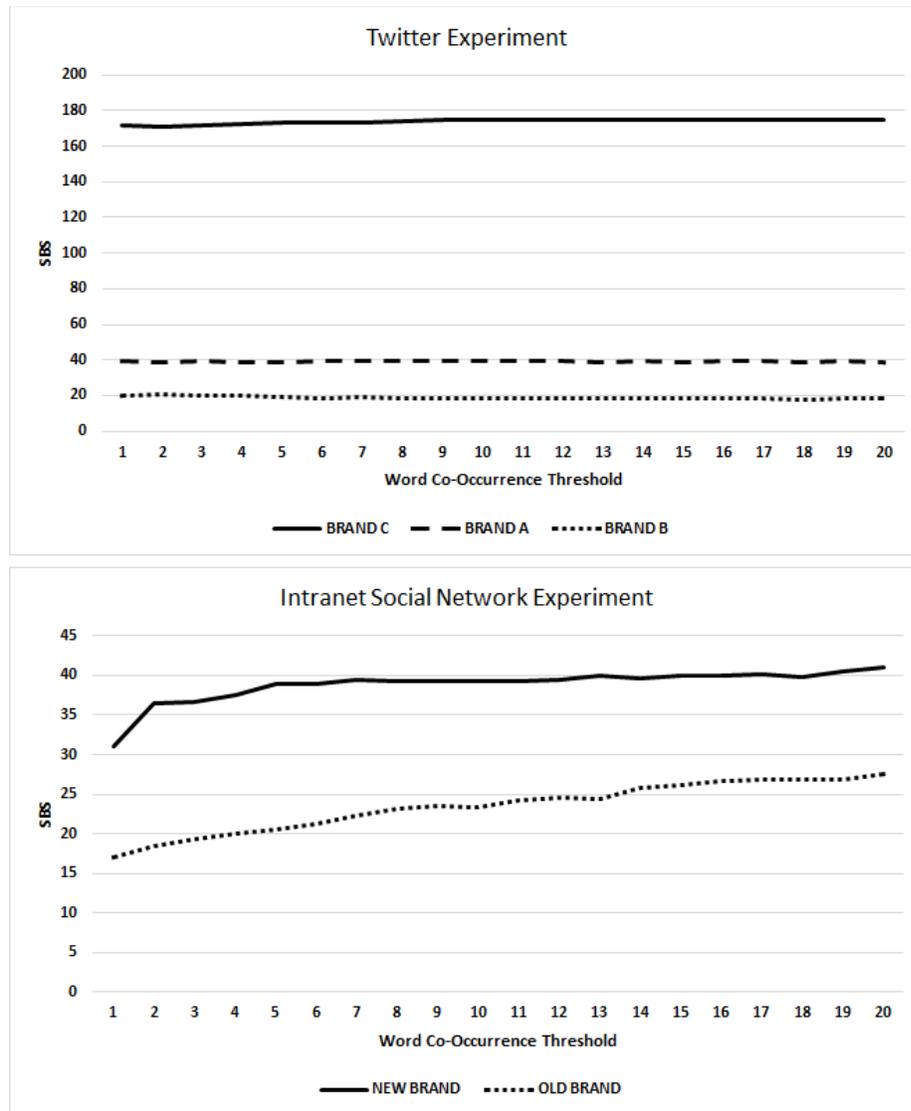

**Figure 5.** Variation of the SBS for different word co-occurrence thresholds.

As the figure shows, even if different word co-occurrence thresholds could determine different absolute values of the SBS, and the gap between brands could partially change, their relative ranking remained the same. Accordingly, one of the most important aspects is to be consistent while replicating the calculation of the SBS for different brands or across different timeframes. In addition, from these first experiments, it seems that the gap between brands tends to be more stable for shorter text documents – where even smaller thresholds cover a larger proportion of total words. These initial results, together with past research, suggest using a threshold of 5 or 7 words (Fronzetti Colladon & Scettri, 2017). However, the analyst should feel free to adjust this parameter, evaluating the co-occurrence range which appears more appropriate in the specific context that is being analyzed. Indeed, the topic is still open for future research as other past studies used very different thresholds, sometimes also reaching 50 words (e.g., Bullinaria & Levy, 2007; Liu & Cong, 2013; Veling & Van Der Weerd, 1999). It is important to remember that the co-occurrence threshold represents the maximum distance of the brand name from the other words in each text document. Therefore, even if choosing very high values is technically possible, such choice could be inappropriate.

This section presented only few examples of possible applications of the SBS in order to provide some hints about its flexibility and potential uses. To extend these preliminary findings is not within the scope of this paper, but will be the objective of future research.

## 5. Discussion and Conclusions

This paper presented the Semantic Brand Score, a new measure of brand importance which combines methods of semantic and social network analysis and can be applied to large text corpora, across products, markets and languages. One advantage of SBS is that it can be used to evaluate the importance of a brand in contexts where consumers, or other stakeholders, can express themselves more spontaneously than when formally interviewed or when in a focus group. The calculation of the SBS does not rely on time-consuming surveys, even if the score can also be calculated on interview transcripts. The SBS can be applied to big data and its measurement is usually more flexible and faster than that of traditional survey-based approaches.

The area of research linking textual analysis with brand management has been explored by a limited number of studies. Compared to other methods and analytical tools for brand management (e.g., Aggarwal et al., 2009; De Vries et al., 2012; Gloor et al., 2009; Yun & Gloor, 2015) the SBS displays several advantages. Firstly, the score has three dimensions of prevalence, diversity and connectivity which are new for a part but also at least partially linked to some pivotal dimensions of well-accepted brand equity models – such as brand awareness and heterogeneity of brand image (Aaker, 1996; Keller, 1993) – and with well-known text statistics such as the study of term frequency and word co-occurrences (Evert, 2005). These three dimensions represent different constructs, offering a more comprehensive final indicator – compared to studies where the final score is limited to the calculation of a single metric or to the analysis of a single construct, such as brand popularity on the web. Secondly, in the calculation of the SBS, social network metrics are applied to word co-occurrence networks, which can be extracted from any text data, making this measure suitable for multiple comparisons – such as the evaluation of brand importance in newspaper articles while considering companies operating in the same business sector. By contrast, some existing analytics are constrained to specific domains (such as Facebook) (e.g., De Vries et al., 2012), or appear to be more appropriate when data is extracted from the web or from online social media (e.g., Gloor, 2017; Gloor et al., 2009), as user interaction, or links among webpages, are an important part of the analysis.

This paper extends the findings of other scholars, who demonstrated that the betweenness centrality of a brand name can be a proxy for its relevance in specific contexts (Fronzetti Colladon & Scettri, 2017; Gloor et al., 2009). The dimension of connectivity represents a new contribution to the research on brand equity, operationalizing a construct which expresses the level of embeddedness of a brand in a discourse (or set of text documents).

This work also extends the research about the possible uses of word co-occurrence networks. In fact, the calculation of the SBS is based on the analysis of these networks, where each word (including the brand) is represented as a node connected to other words by weighted edges. This representation can be very useful for those who would like to study brand image and consider the most important associations of the brand with other words within the text. Ultimately, it can be a starting point for both the categorization of the textual brand associations and the study of their valence, thus supplementing the information provided by the SBS.

The SBS has the additional advantage of being flexible as it can be calculated for any set of words, without being limited to brands. One could, for example, study an intranet social network and the evolution of the SBS for a company's core values, as perceived and discussed by the employees. The proposed metric can also be used for comparative purposes to assess the positioning of a brand with respect to its competitors. Brand managers could use the SBS, for example, to compare the importance of their brand in newspaper articles, on social media, in the perception of internal employees or customers. In this sense, insights coming from the temporal evolution of the SBS, in settings which can be both internal and external to a company, can support decision-making processes. In general, the knowledge of the value of a brand can significantly influence many aspects of a company's life, such as the recruiting process (Franca & Pahor, 2012) or the decisions made about advertising campaigns or marketing strategies (Keller, 2009). The SBS can also be used as a starting point for the evaluation of consumers' feedback on social media, as a mean to support customer relationship management (Malthouse et al., 2013).

Even if the measure is new, there is evidence to suggest that the dimensions of the SBS could be useful for financial forecasting purposes (see Section 4) and provide results which are consistent with past research (Fronzetti Colladon & Scettri, 2017). Indeed, the SBS opens a new research branch about its possible applications. As brand equity proved to be connected with sales and other indicators of firm performance (e.g., H. B. Kim & Kim, 2005), it is not excluded that the SBS and the sentiment of brand co-occurrences, measured for example on social media, could be of help in making better predictions. For example, the SBS could be used by hotel managers to measure their positioning with respect to their competitors on major online platforms such as TripAdvisor or Facebook, in order to forecast sales and plan social media marketing actions to improve brand equity (Yazdanparast, Joseph, & Muniz, 2016). The hotel industry, indeed, is another reality where brand equity is connected to financial performance (H. B. Kim, Gon Kim, & An, 2003). Outside the business world, the SBS could serve to assess the importance of personal brands, such as the name of political candidates, to test whether this could support the prediction of political outcomes. Indeed, past research suggested that brand image of political candidates can influence electoral decisions (Guzmán & Sierra, 2009). A network representation of brand co-occurrences could provide insight in this context and be utilized to infer possible brand associations and support the evaluation of brand image.

It is important to notice that the SBS is not an overall measure of brand equity and its value can change depending on the context that is being analyzed. Therefore, the final score should always be presented together with the characteristics of the domain which is being considered, and possibly with an evaluation of the positivity or negativity of brand co-occurrences. Consistently, a company brand might have a very high SBS when looking at the discourse taking place on major social media websites and, by contrast, have a low SBS if considering the newspapers articles that relate to the business sector in which the company operates.

The idea of the SBS has some limitations. Firstly, as already mentioned in Section 3, this index is sample dependent, i.e. it changes while choosing different text corpora. This is for a part totally normal and common to many other indicators – as it would happen for instance with the Net Promoter Score when changing the sample of customers who are interviewed (Reichheld, 2003). Accordingly, the analyst should be careful while comparing scores obtained in different contexts (e.g. conversations on Twitter and Facebook). Sometimes it is more fruitful to discuss the relative score distances among a set of selected brands or the SBSs evolution over time, instead of just focusing on absolute numbers. A second limitation comes from choosing to calculate the SBS as the sum of prevalence, diversity and connectivity, in a way that attributes the same importance to each individual score. Future research could discuss the possibility of using different coefficients, to value one dimension over the others, and explore other aggregation and normalization methods. A simple way to mitigate this limitation is to always read and comment the final value of the SBS together with the values of its components. In his preliminary experiment, for example, the author found that connectivity had more informative power for the prediction of stock prices than the other two dimensions of the SBS.

An additional limitation derives from the fact that some small and medium enterprises might lack adequate infrastructure and skilled personnel to replicate the methodology presented in this paper. To address this issue, it is the author's intent to develop a multi-platform software for the calculation of the Semantic Brand Score, to support both business analysts and company managers. Another common problem might arise when datasets are very big and handling them efficiently – to also reduce computational times – can be challenging or require significant IT resources. This is a limitation that SMEs could at least solve partially by using distributed computing and cloud solutions (Jacobs, 2009).

Lastly, even if the algorithms used for the calculation of the SBS are applicable to large sets of documents, another problem could be the methodology used to collect and integrate big text data from multiple sources (Ziegler & Dittrich, 2004). A detailed answer is not within the scope of this research and cannot be generalized, as it varies depending on the research setting. Moreover, when datasets become very big, the SBS undergoes the same challenging issues of all data driven models (Wu, Zhu, Wu, & Ding, 2014). Beyond certain limits, data aggregation and integration can be difficult or resource intensive. To help in this direction, some tools are already available on the market: for example, Condor is a software meant to handle big network data; it has proved to be very useful for the collection of text documents from the web (being able to crawl Twitter, Facebook, Wikipedia, blogs and other webpages from Google). Condor has also a built-in function to extract content from email networks (Gloor, 2017). Other applications might require a specific crawler or access to premade databases, such as GDELT (Leetaru & Schrodt, 2013). The SBS can be calculated on datasets which only contain two fields: one with the content of each text document and one with a timestamp (useful if the analysis is longitudinal). As suggested in Section 3, the preprocessing of text documents – for example to remove special characters and stop-words – can be implemented using the functions included in the package NLTK, developed for the Python programming language (Perkins, 2014). To the same purpose, the use of other programming languages or tools – such as R (Roberts et al., 2015) or the software Context (Diesner, 2014) – is also possible. The software program, which is under development by the author, will include all the functions required to preprocess text documents and calculate the SBS, but has not been conceived for data collection purposes. The calculation is made on pre-collected data.

Future research could extend this work by better linking the constructs of brand image (Keller, 1993) and associations with the textual co-occurrences of a brand. The co-occurrence network could be examined as a novel starting point to construct brand maps (John, Loken, Kim, & Monga, 2006). Additionally, future studies could consider better integration of sentiment analysis with the SBS and propose new ways to assess the sentiment of the words which are linked to a brand. Scholars could also explore other construction techniques for word co-occurrence networks – for example considering punctuation or more appropriate sentence boundary disambiguation algorithms (Palmer & Hearst, 1997) – or better investigate the effects of changing the threshold for the sliding window of the co-occurrence range, which does not affect prevalence but can change the values of diversity and connectivity. Lastly, future research could combine social network and semantic analysis to discover possible new dimensions to be added to the SBS.

The Semantic Brand Score is a new indicator which can be highly informative for brand managers and scholars; it should not be intended as a surrogate of existing brand equity models or measurement techniques. This metric has an important connection with past research and models but, at the same time, it introduces new constructs which express brand importance in text data and can integrate brand equity research with methods from social network and semantic analysis. Overall, the SBS can be used to support decision-making processes within companies. In the era of big data, this indicator offers an important contribution to the strategic management of brands as long-term assets.